\documentclass[runningheads]{llncs}
\usepackage[T1]{fontenc}
\usepackage{graphicx}
\usepackage{amssymb}
\usepackage{xcolor}
\usepackage{colortbl}
\usepackage{booktabs}
\usepackage{multirow, makecell}
\usepackage[colorlinks,linkcolor=red,anchorcolor=blue,citecolor=green]{hyperref}
\usepackage{orcidlink}
\usepackage{amsmath,mathtools}
\usepackage{algorithm}
\usepackage{algorithmic}
\usepackage[switch]{lineno}
\usepackage{bbding, pifont}
\usepackage{soul}
\usepackage{url}

\begin{document}
%
% replace DEYOLO to CDFETOLO, esay pronunciation 
\title{DEYOLO: Dual-Feature-Enhancement YOLO for Cross-Modality Object Detection}
\titlerunning{Dual-Feature-Enhancement YOLO}
% If the paper title is too long for the running head, you can set
% an abbreviated paper title here
%

\author{Yishuo Chen\inst{1}~\orcidlink{0009-0009-6853-5734} \and
 Boran Wang\inst{1(}\Envelope \inst{)} ~\orcidlink{0009-0002-7626-1154} \and 
 Xinyu Guo\inst{1}~\orcidlink{0009-0008-6472-1719} \and
 Wenbin Zhu\inst{1}~\orcidlink{0000-0002-2416-9332} \and
 Jiasheng He\inst{1}~\orcidlink{0009-0005-3224-1284} \and
 Xiaobin Liu\inst{1,2} \and
 Jing Yuan\inst{1,2}~\orcidlink{0000-0001-5495-684X}}
 
\authorrunning{Yishuo et al.}
% First names are abbreviated in the running head.
% If there are more than two authors, 'et al.' is used.
%
% \institute{Princeton University, Princeton NJ 08544, USA \and
% Springer Heidelberg, Tiergartenstr. 17, 69121 Heidelberg, Germany
% \email{lncs@springer.com}\\
% \url{http://www.springer.com/gp/computer-science/lncs} \and
% ABC Institute, Rupert-Karls-University Heidelberg, Heidelberg, Germany\\
% \email{\{abc,lncs\}@uni-heidelberg.de}}

\institute{ College of Artificial Intelligence, Nankai University, Tianjin 300350, China \and Engineering Research Center of Trusted Behavior Intelligence, Ministry of Education, Nankai University, Tianjin 300350, China  \\
\Envelope \ \ {\href{mailto:wangbr1025@gmail.com}{\textcolor{black}{wangbr1025@gmail.com}}}
}

%\and Tianjin Key Laboratory of Intelligence Robotics, Nankai University, Tianjin 300350, China 
\maketitle              % typeset the header of the contribution

\begin{abstract}
Object detection in poor-illumination environments is a challenging task as objects are usually not clearly visible in RGB images. 
% Existing methods for this task commonly employ RGB images only, hindering accurately detecting objects due to the limited wavelength range of visible cameras. 
As infrared images provide additional clear edge information that complements RGB images, fusing RGB and infrared images has potential to enhance the detection ability in poor-illumination environments.
% For object detection task in poor-illumination environments, existing methods using RGB images only are not suitable due to the limited wavelength range of the visible camera. Therefore, the visible-infrared image fusion provides an alternative solution. 
However, existing works involving both visible and infrared images only focus on image fusion, instead of object detection. Moreover, they directly fuse the two kinds of image modalities, which ignores the mutual interference between them. 
To fuse the two modalities to maximize the advantages of cross-modality, we design a dual-enhancement-based cross-modality object detection network DEYOLO, in which semantic-spatial cross-modality and novel bi-directional decoupled focus modules are designed to achieve the detection-centered mutual enhancement of RGB-infrared (RGB-IR). Specifically, a dual semantic enhancing channel weight assignment module (DECA) and a dual spatial enhancing pixel weight assignment module (DEPA) are firstly proposed to aggregate cross-modality information in the feature space to improve the feature representation ability, such that feature fusion can aim at the object detection task. Meanwhile, a dual-enhancement mechanism, including enhancements for two-modality fusion and single modality, is designed in both DECA and DEPA to reduce interference between the two kinds of image modalities. Then, a novel bi-directional decoupled focus is developed to enlarge the receptive field of the backbone network in different directions, which improves the representation quality of DEYOLO. 
Extensive experiments on M$^3$FD and LLVIP show that our approach outperforms SOTA object detection algorithms by a clear margin. Our code is available at \href{https://github.com/chips96/DEYOLO}{https://github.com/chips96/DEYOLO}.

\end{abstract}

\section{Introduction}\label{section:Introduction}

As a fundamental task of computer vision, object detection in complex scenes still encounters various challenges. Due to the limited wavelength range of visible light, it is difficult to obtain object information in complex environments with poor illumination (\emph{e.g.} heavy smoke). To address this problem, infrared information has been widely introduced. However, due to the low quality of infrared images, it is hard to extract useful texture and color information for general detectors from infrared images. Thus, it is difficult for them to support the detection task alone.

In contrast, utilizing the complementary information in the cross-modality of visible-infrared images can improve the performance in object detection. The commonly used methods adopt fusion-and-detection strategies, which means the image fusion network uses the object detection results as the validation metric. However, the fusion-and-detection methods have several deficiencies. Firstly, fusion of two-modality images does not focus on object detection tasks. Secondly, their redundant model structures (\emph{e.g.} two separate models for fusion and detection, respectively) cause increased training cost as well. Thirdly, although being rich in structure information, infrared (IR) images have a drawback of missing texture. Thus, fusion models usually focus on enriching the texture information while eliminating the complex brightness information of the object. On the contrary, they seldom take the mutual interference between the two modal images into account. \emph{e.g.} infrared images maybe offset the visible imaging quality in fusion process. Only direct image pair fusion without cross-modality enhancement is not sufficient to improve the object detection performance.

Most existing RGB-IR detection models either construct a four-channel input or maintain RGB and infrared images in two separate branches, merging their features downstream. These multi-modality information fusion strategies enhance detection performance to some extent. However, we believe that the interaction between the two modalities is insufficient in these methods. There is a clear boundary between the processing of single-modality images and the feature fusion, resulting in insufficient utilization of cross-modality information. Furthermore, they lack compound interactions at the channel and spatial dimensions, overlooking the potential relationship between semantic and structural information.

\begin{figure}[h]
    \centering
    \includegraphics[width=\textwidth]{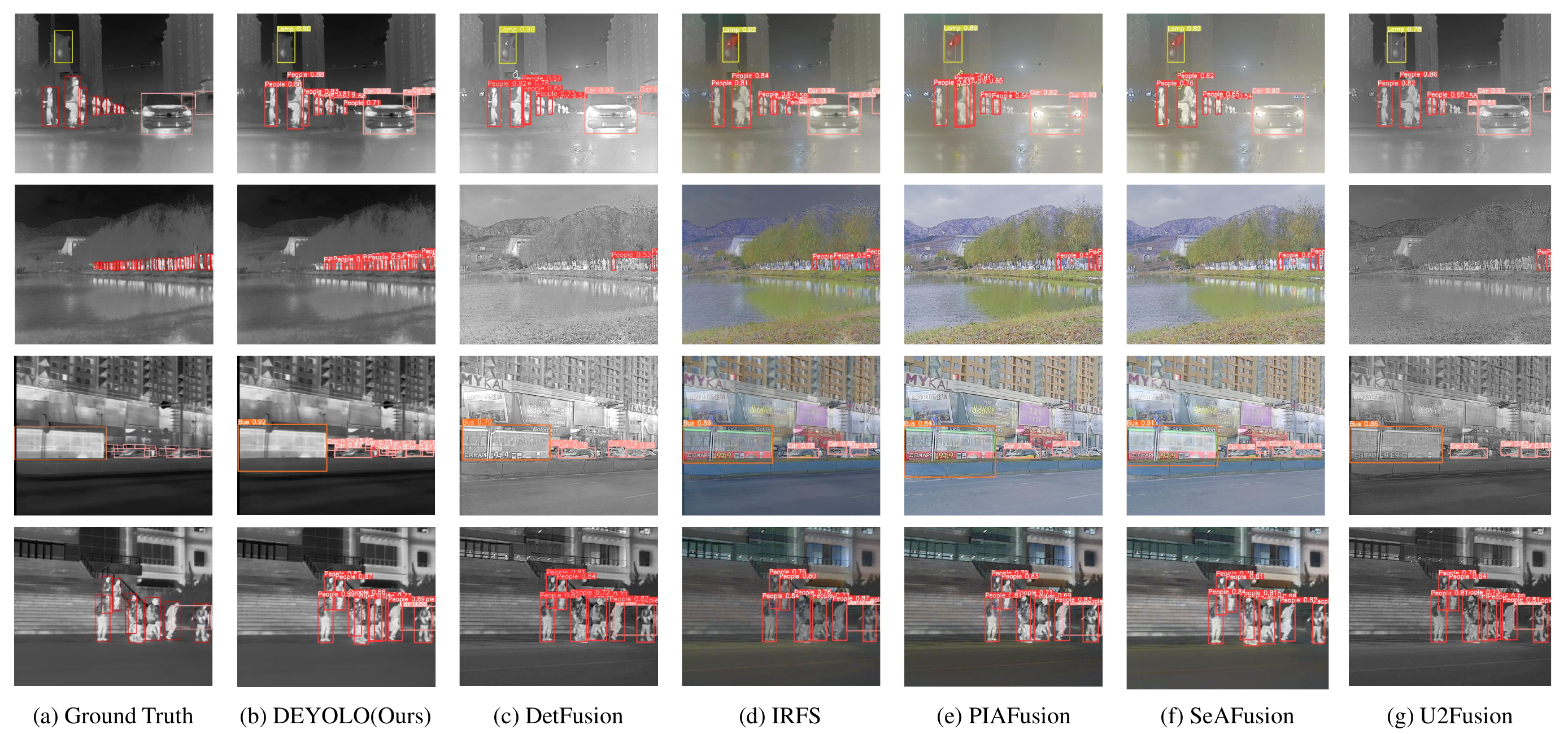}
    \vspace{-0.5cm}
    \caption{Detection results of different methods.}
    \label{fig:comparison}
\end{figure}

To this end, we propose a cross-modality feature fusion approach to dually enhance the feature map of visual and infrared images for detection tasks. This enhancement strategy is able to guide the fusion process of two-modality features from different scales to ensure the integrity of feature information and optimal information extraction. Aiming at object detection, DECA and DEPA are designed to enrich semantic and structure information contained in the feature maps respectively. Moreover, for the purpose of highlighting the modality-specific characteristics, we insert a novel bi-directional decoupled focus in the backbone. It improves the receptive field in the feature extraction stage of DEYOLO multi-directionally, yielding better results. Fig.~\ref{fig:comparison} shows the detection results by DEYOLO and DetFusion~\cite{sun:DetFusion}, IRFS~\cite{WANG2023101828}, PIAFuse~\cite{tang:PiaFusion},SeaFusion~\cite{tang:SeAFusion} U2Fsuion~\cite{xu:U2Fusion}. It can be observed that the proposed DEYOLO achieve better detection results. The contributions of this work are three-fold:
\begin{enumerate}
	\item We propose the DEYOLO based on YOLOv8~\cite{YoloV8}, which performs cross-modality feature fusion between the backbone and the detection heads. Different from other fusion methods which directly fuse  two-modality images, we fuse two-modality information in feature space and focus on object detection tasks.
	\item We propose two modules DECA and DEPA utilizing dual-enhancement mechanism. They reduce interference between two kinds of modalities and achieve semantic and spatial information enhancement by redistributing the weights of channels and pixels.
	\item To make the features extracted by the backbone more adaptive to our  dual-enhancement mechanism, we design the bi-direction decoupled focus. It downsamples shallow feature maps in different directions, increasing the receptive fields without losing surrounding information.
\end{enumerate}

\section{Related Work}

In this section, we review the commonly used single-modality object detection algorithms first. Then, some recent visible and infrared image fusion methods are introduced.

\subsection{Single-Modality Object Detection}

Recently, deep neural networks have been proposed to improve accuracy in object detection tasks, including CNN and its variants, \emph{e.g.}  Sparse R-CNN~\cite{sun:sparseRCNN}, CenterNet2~\cite{zhou:CenterNet2} and the YOLO series~\cite{redmon:YOLOV1,bochkovskiy:YOLOV4,wang:YOLOV7}, as well as Transformer-based models, \emph{e.g.} DETR~\cite{carion:DETR} and Swin Transformer~\cite{liu:Swin}. Although the outstanding performance can be achieved by these models, they all merely utilize information from single-modality images. In addition, these models heavily rely on the texture of the image, which hinders their detection capabilities for infrared images.

To handle infrared object detection problems, researchers are continuously introducing different network structures and mechanisms. ALCNet~\cite{dai:ACLNet} uses backbone to extract the high-level semantic features of the image and a model-driven encoder to learn the local contrast features. ISTDU-Net~\cite{9674870} effectively integrates the encoding and decoding stages and facilitates the transfer of information through hopping connections. This structure is able to increase the receptive field while maintaining a high resolution. IRSTD-GAN~\cite{zhao:IRSTD-GAN} treats infrared targets as a special kind of noise. It can predict infrared small targets from the input image based on the data distribution and hierarchical features learned by the GAN. These models only take infrared images into account without extracting information from visible images.

The above single-modality methods are not well suitable for object detection under complex illumination conditions. In contrast, two-modality fusion can extract complementary information from both visible and infrared images, and thus has less over-dependence on texture information.

\begin{figure*}[t]
	\centering
	\includegraphics[width=.9\linewidth]{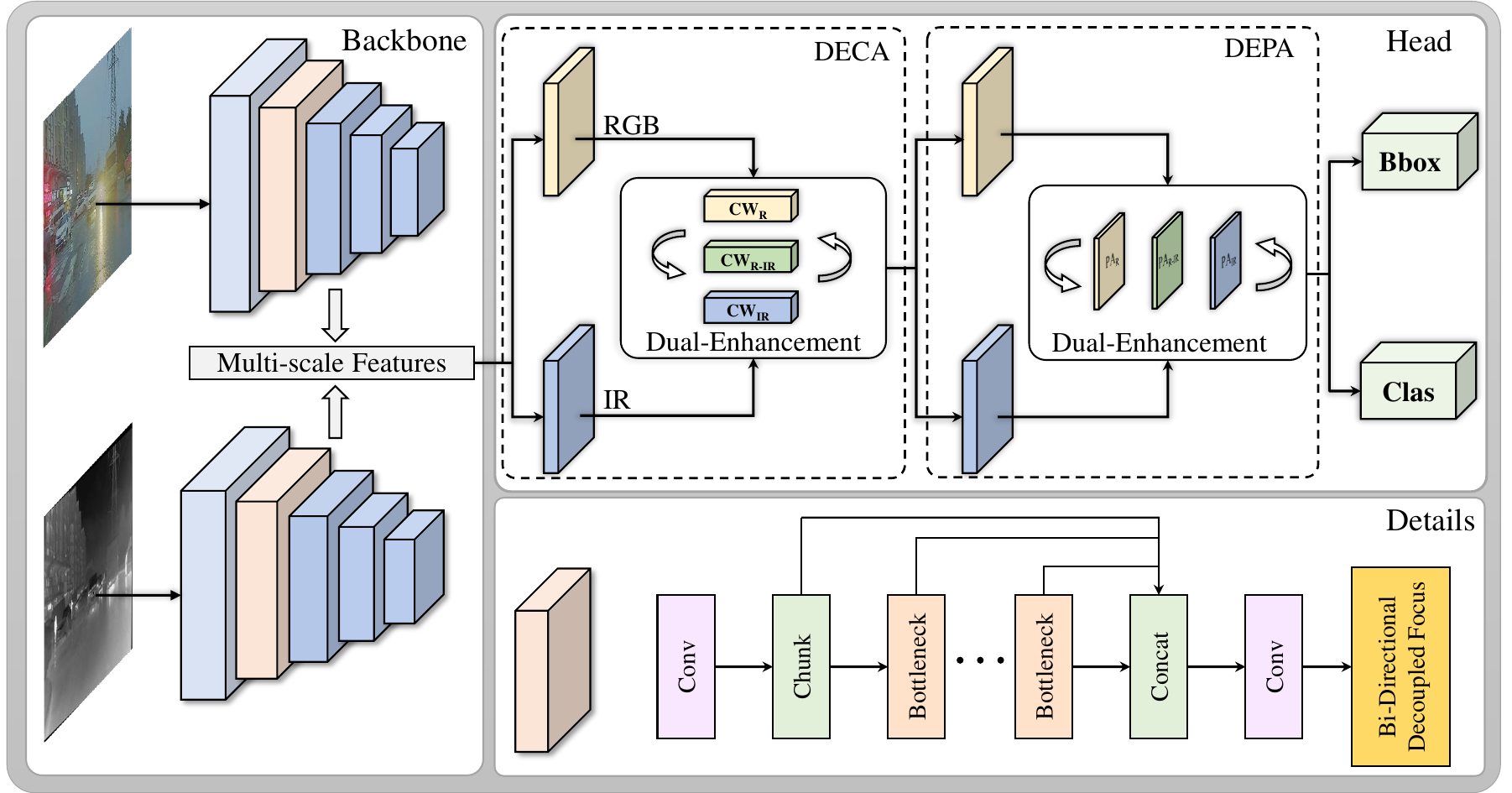}
 % \vspace{-0.5cm}	
	\caption{The framework of the proposed DEYOLO. We incorporate dual-context collaborative enhancement modules (DECA and DEPA) within the feature extraction streams dedicated to each detection head in order to refine the single-modality features and fuse multi-modality representations. Concurrently, the Bi-direction Decoupled Focus is inserted in the early layers of the YOLOv8 backbone to expand the network's receptive fields.}
	\label{network}
\end{figure*}

\subsection{Fusion-and-Detection Methods}\label{section:Fusion-and-detection}
Considering that infrared images are less vulnerable to poor lighting conditions, various visible and infrared image fusion methods have been proposed.

U2Fusion~\cite{xu:U2Fusion} is an unsupervised end-to-end image fusion network that can solve different fusion problems. It uses feature extraction and information measurement to automatically estimate the importance of the corresponding source images and proposes adaptive information preservation degree. PIAFusion~\cite{tang:PiaFusion} takes the illumination factor into account using an illumination-aware loss. SwinFusion~\cite{ma:SwinFusion} involves  fusion units based on self-attention~\cite{vaswani2017attention} and  cross-attention, in order to mine long dependencies within the same domain and across domains. CDDFuse~\cite{Zhao:CDDFuse} introduces a Transformer-CNN extractor and succeeds in decomposing desirable modality-specific and modality-shared features. After the fusion process, the obtained image are fed to a separate model to detect objects.

Although these models can produce convincing results that preserve the adaptive similarity between the fusion result and source images, they don't  directly aim at the object detection task. Another drawback is that there may exist conflicts in the fusion results (\emph{e.g.} the textureless patches of infrared images ruin the originally texture-rich ones of visible images), which is harmful to detection accuracy. In contrast, DEYOLO only focuses on object detection and the newly designed dual-enhancement mechanism can tackle the conflict problem.

\section{Method}

As shown in Fig.\ref{network}, to process the multi-scale features extracted from the two-modality images, we add newly designed modules DECAs and DEPAs (Fig.\ref{DECA and DEPA}) between the backbone and the necks of the YOLOv8~\cite{YoloV8} model. Through a specific dual-enhancement mechanism, the fusion of semantic and spatial information makes two-modality features more harmonious. Meanwhile, for the backbone network, to better extract and retain the useful features of both modalities of images, we propose a novel bi-directional decoupled focus strategy. It increases the receptive field of the backbone in different orientations and ensures no leakage of origin information.

\subsection{DECA: Dual Semantic Enhancing Channel Weight Assignment Module}

The dual enhancement mechanism here refers to the enhancement for two-modality fusion result with single-modality information between the channels and further enhancement for single modality with complementary information from two-modality fusion. Therefore, DECA is able to emphasize the semantic information by distributing weights according to the importance of each channel.

The first enhancement aims to use the single-modality feature to improve the two-modality fusion results of both RGB-IR features, which may contain conflicts. Let $\boldsymbol{F}_{V_0} \in \mathbb{R}^{b \times c \times h \times w}$ and $\boldsymbol{F}_{IR_0} \in \mathbb{R}^{b \times c \times h \times w}$ be the feature maps of visible and infrared images calculated by the backbone, respectively. At first, to get the comprehensive information of RGB and IR images, we concatenate the two features along the channel dimension. Then, a convolution operation will make the combined feature map change to the previous size, filtering the redundant information. As a result, the mixed feature map $\boldsymbol{F}_{Mix_0} \in \mathbb{R}^{b \times c \times h \times w}$ is obtained:
\begin{equation}
	\boldsymbol{F}_{Mix_0} = conv(concat(\boldsymbol{F}_{V_0},\boldsymbol{F}_{IR_0}))
	\label{DECA first enhance1}
\end{equation}

Next, we propose a novel weight-encoding method through convolution. An encoder is designed to squeeze $\boldsymbol{F}_{Mix_0}$ in the spatial dimension progressively to the size of $\mathbb{R}^{b \times c \times 1 \times 1}$:
\begin{equation}
	\boldsymbol{W}_{Mix_0} = CMWE(\boldsymbol{F}_{Mix_0}) \in \mathbb{R}^{b \times c \times 1 \times 1}
	\label{DECA first enhance2}
\end{equation}

\begin{figure*}[t]
	\centering
	\includegraphics[width=1\linewidth]{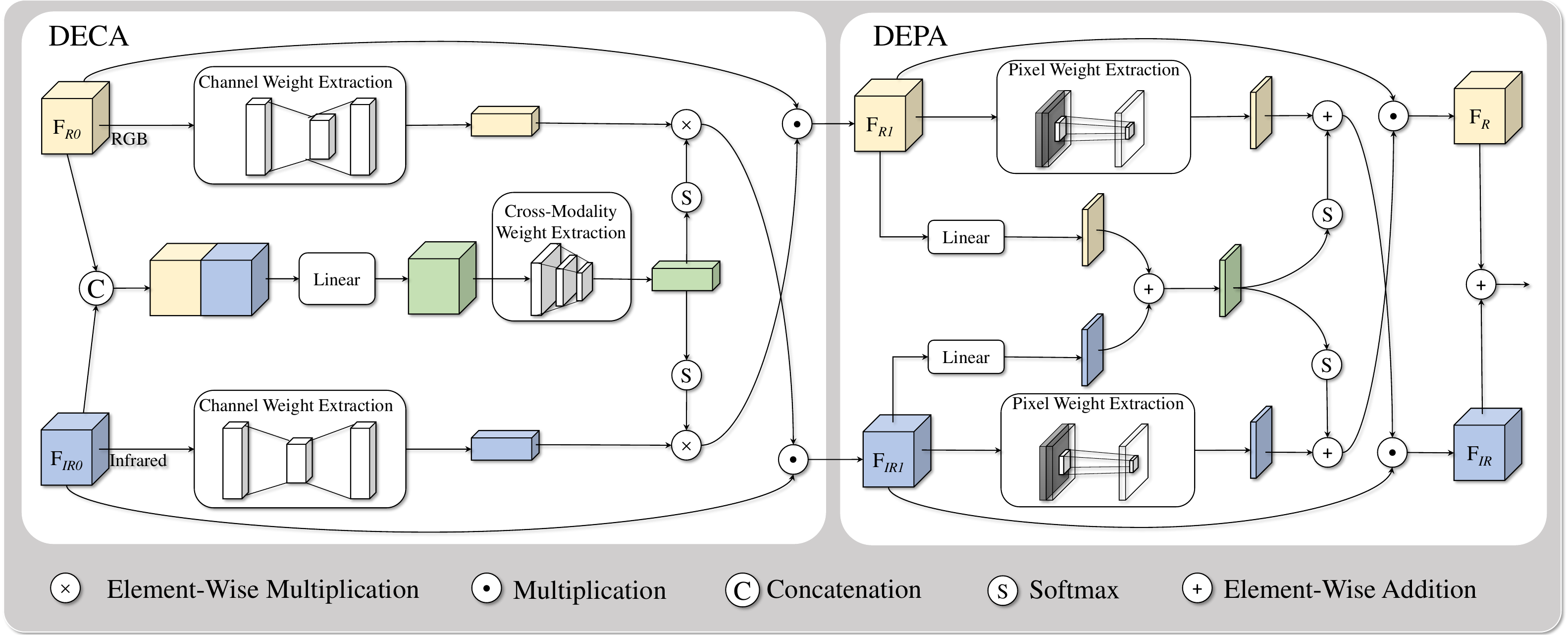}
\vspace{-0.5cm}	
 \caption{The concrete structure of DECA and DEPA. These modules utilize both single-modality and cross-modality information through a dual enhancement mechanism. DECA enhances the cross-modality fusion results by leveraging dependencies between channels within each modality and outcomes are then used to reinforce the original single-modal features, highlighting more discriminative channels. Similarly, DEPA is able to learn dependency structures within and across modalities to produce enhanced multi-modality representations with stronger positional awareness.}
	\label{DECA and DEPA}
\end{figure*}

\noindent
where $CMWE(\cdot)$ refers to the cross-modality weight extraction operation in Fig.\ref{DECA and DEPA}.

On the other hand, we need to acquire the specific feature of each modality. The SE block~\cite{hu:SEattention} explicitly models the interdependencies between the channels of its convolutional features for improving the quality of the feature map representation. Motivated by this idea, we feed this structure with visible and infrared images to get the feature blocks of size $\mathbb{R}^{b \times c \times 1 \times 1}$, which represents the weight values of different channels:
\begin{equation}
\left\{\begin{aligned}
	&\boldsymbol{W}_{V_0} = CWE(F_{V_0}) \in \mathbb{R}^{b \times c \times 1 \times 1} \\
	&\boldsymbol{W}_{IR_0} = CWE(F_{IR_0}) \in \mathbb{R}^{b \times c \times 1 \times 1}
	\label{DECA first enhance3}
\end{aligned}\right.
\end{equation}
where $CWE(\cdot)$ refers to the channel weight extraction block in Fig.\ref{DECA and DEPA}. $\boldsymbol{W}_{V_0}$ and $\boldsymbol{W}_{IR_0}$ can enhance the mixed feature of the two modalities by element-wise multiplication to redistribute weights, which is able to highlight significant channels:
\begin{equation}
\left\{\begin{aligned}
	&\boldsymbol{W}_{enV_0} = \boldsymbol{W}_{V_0} \otimes softmax(\boldsymbol{W}_{Mix_0}) \\
	&\boldsymbol{W}_{enIR_0} = \boldsymbol{W}_{IR_0} \otimes softmax(\boldsymbol{W}_{Mix_0}) 
	\label{DECA first enhance4}
\end{aligned}\right.
\end{equation}

For the second enhancement, we attempt to make each feature map of RGB and IR fully utilize the respective advantages of another modality. To this end, $\boldsymbol{F}_{V_0}$ and $\boldsymbol{F}_{IR_0}$ will multiply the corresponding feature weights acquired in the first enhancement to get semantic and textural information from another modality:
\begin{equation}
\left\{\begin{aligned}
	&\boldsymbol{F}_{IR_1} = \boldsymbol{F}_{IR_0} \odot \boldsymbol{W}_{enV_0} \\
	&\boldsymbol{F}_{V_1} = \boldsymbol{F}_{V_0} \odot \boldsymbol{W}_{enIR_0}
	\label{DECA second enhance}
\end{aligned}\right.
\end{equation}
where $\odot$ is multiplication in channel dimension. The enhancement results $\boldsymbol{F}_{V_1} \in \mathbb{R}^{b \times c \times w \times h}$ and $\boldsymbol{F}_{IR_1} \in \mathbb{R}^{b \times c \times w \times h}$ will pass through the DEPA described below.

\subsection{DEPA: Dual Spatial Enhancing Pixel Weight Assignment Module}
Similar with DECA, DEPA adopts the dual enhancement mechanism as well.  Re-encoded in the spatial dimension, DEPA emphasizes important pixel positions while minimizing the irrelevant ones.

Specifically, to obtain the mixed feature including global information, we perform a shape transformation for the two feature maps  $\boldsymbol{F}_{V_1}$ and $\boldsymbol{F}_{IR_1}$ using convolution. Then, an element-wise multiplication is applied on the result of each other: 
\begin{equation}
	\boldsymbol{W}_{Mix_1} = conv(\boldsymbol{F}_{V_1}) \otimes conv(\boldsymbol{F}_{IR_1})
	\label{DEPA first enhance1}
\end{equation}
Afterwards, a softmax operation is performed on $\boldsymbol{W}_{Mix_1}$. In order to fully obtain the feature specific to each modality in spatial dimension, we maintain the differences in spatial information learned by different convolutional kernel sizes. 
\begin{equation}
\left\{\begin{aligned}
	&\boldsymbol{W}_{IR_1temp} = concat\left( conv_{1}(\boldsymbol{F}_{IR_1}), conv_{2}(\boldsymbol{F}_{IR_1}) \right) \\
	&\boldsymbol{W}_{V_1temp} = concat\left( conv_{1}(\boldsymbol{F}_{V_1}), conv_{2}(\boldsymbol{F}_{V_1}) \right)
	\label{DEPA first enhance2}
\end{aligned}\right.
\end{equation}
In Eq.(\ref{DEPA first enhance2}), two convolution operations are used to extract the pixel weights from distinct scales. By concatenating them in the channel dimension, we can obtain $\boldsymbol{W}_{IR_1} \in \mathbb{R}^{b \times 2\times w \times h}$ and $\boldsymbol{W}_{V_1} \in \mathbb{R}^{b \times 2\times w \times h}$. Then, we compress the feature by reducing the number of channels by half and obtain $\boldsymbol{W}_{IR_1} \in \mathbb{R}^{b \times 1\times w \times h}$ and $\boldsymbol{W}_{V_1} \in \mathbb{R}^{b \times 1\times w \times h}$. The element-wise multiplication by the softmaxed $\boldsymbol{F}_{Mix_1}$ is applied on $\boldsymbol{W}_{IR_1}$ and $\boldsymbol{W}_{V_1}$:
\begin{equation}
\left\{\begin{aligned}
	&\boldsymbol{W}_{enIR_1} = \boldsymbol{W}_{IR_1} \otimes softmax(\boldsymbol{F}_{Mix_1}) \\
	&\boldsymbol{W}_{enV_1} = \boldsymbol{W}_{V_1} \otimes softmax(\boldsymbol{F}_{Mix_1})
	\label{DEPA first enhance3}
\end{aligned}\right.
\end{equation}

The second enhancement is implemented by an element-wise multiplication operation between the input feature maps and the results of first enhancement:
\begin{equation}
\left\{\begin{aligned}
	&\boldsymbol{F}_{IR} = \boldsymbol{F}_{IR_1} \odot \boldsymbol{W}_{enV_1} \\
	&\boldsymbol{F}_{V} = \boldsymbol{F}_{V_1} \odot \boldsymbol{W}_{enIR_1}
	\label{DECA second enhance}
\end{aligned}\right.
\end{equation}
Eq.(\ref{DECA second enhance}) aims to extract structural feature from another modality in spatial dimension. In the end, we do element-wise addition on $\boldsymbol{F}_{IR}$ and $\boldsymbol{F}_{V}$ for the object detection.

\subsection{Bi-direction Decoupled Focus}

\begin{figure}[h]
	\centering
	\includegraphics[scale=.3]{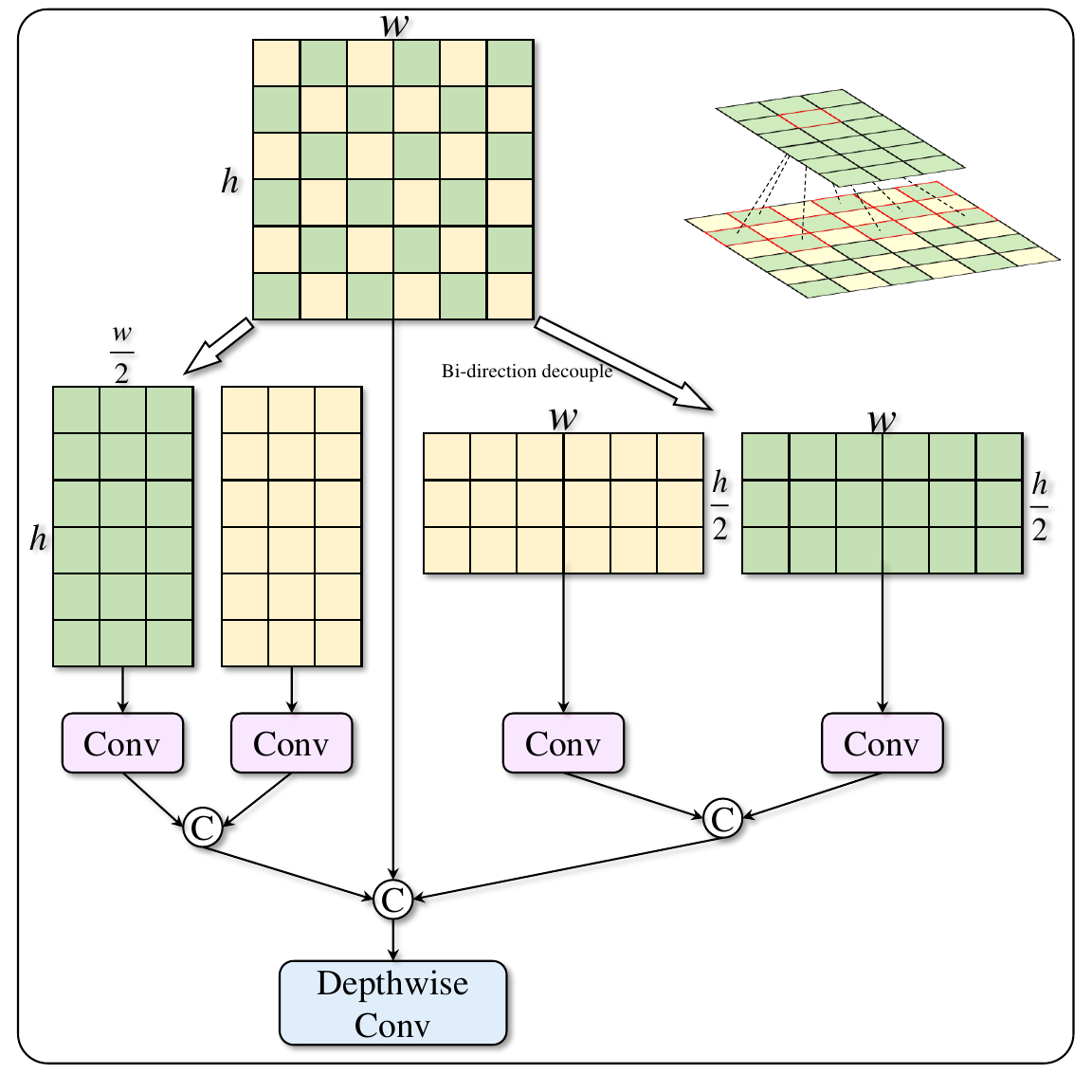}
\vspace{-0.3cm}
 \caption{Bi-direction decoupled focus.}
	\label{focus}
\end{figure}

In this subsection, we tend to improve the performance of object detection from the perspective of the single modality. In order to enhance the capability of extracting targets, the bi-direction decoupled focus is designed to enlarge the receptive field of the backbone in DEYOLO while minimizing the loss of surrounding pixels.

The focus block in YOLOv5~\cite{Jocher_YOLOv5_by_Ultralytics_2020} is a slicing operation, which is improved from the passthrough layer in YOLOv2~\cite{redmon:YOLOV2}. This specific operation gets a pixel in an image with an interval by one pixel  and thus can provide a two-fold downsampled feature map without an information loss.

Inspired by this downsampling method, we design bi-direction decoupled focus to retain the information adequately in multi-directions. Specifically, we adopt two specific sampling and encoding rules implemented horizontally and vertically. As shown in Fig.\ref{focus}, we divide the pixels into two groups for convolution. Each group focuses on the adjacent and remote pixels at the same time. Finally, we concatenate the original feature map in the channel dimension and make it go through a depth-wise convolution~\cite{chollet:Xception}  layer.

\section{Experiments}
% Summerize what we have done 
\subsection{Datasets}
Since infrared images are obtained by measuring the heat radiation emitted from objects, they are susceptible to noises in the environment. In fact, only a small number of high-quality datasets composed of infrared and visible images are available, such as TNO~\cite{toet:TNO} and 
RoadScene~\cite{xu2020aaai}.However, these datasets often aim at infrared and visible image fusion tasks, rather than object detection, thus the labels for object detection are absent. The FLIR\cite{FLIR} dataset provides annotations for object detection but it lacks pixel-level alignment. Therefore, we choose the public datasets M$^3$FD~\cite{liu:Tardal}, LLVIP~\cite{jia:LLVIP}, and KAIST\cite{KAIST} which are pixel-wise aligned for infrared-visible image pairs and contain annotations for object detection. Among these, the M$^3$FD dataset comprises 4,200 image pairs, totaling 8,400 images. The LLVIP dataset includes 16,836 image pairs, amounting to a total of 33,672 images. Considering the original KAIST dataset contains noisy annotations, we use a cleaned version of the training set (7,601 examples) and the testing set (2,252 examples).
% However, these datasets often aim at infrared and visible image fusion, rather than object detection, where either the labels for object detection are absent, or the pixel-wise registration is not provided, which makes it difficult to meet the requirement of cross-modality image pairs for the problem considered in this paper.

\subsection{Implementation details}
In this subsection, two sets of experiments are conducted to verify the effectiveness of DEYOLO. One is the comparison with the SOTA single-modality object detection algorithms and the other is the comparison with the  fusion-and-detection algorithms. When training single-modality detection algorithms, we use infrared and visible images to train the model, respectively. For the sake of experimental fairness, we also combined the visible and infrared images from the datasets to serve as the training set of these detector. For the fusion-and-detection algorithms, the pre-trained image fusion models for cross-modality fusion are adopted in the comparison algorithms, and then the fused images are further used to train YOLOv8~\cite{YoloV8}. 
The training is performed on eight NVIDIA RTX 4090 GPUs. The number of epochs for training is 800, the batch size is 64, the initial and final learning rates are $1 \times 10^{-2}$ and $1 \times 10^{-4}$, respectively. And, we evaluate our method on the validation set and use the mean average precision (mAP) with the IoU threshold of 0.5 and Log Average Miss Rate (LAMR) as the evaluation metric.

\subsection{Ablation Studies}
To validate the impact of the key components in DEYOLO, we conducted a number of experiments on the M$^3$FD~\cite{liu:Tardal} dataset to investigate how they affect our final performance.

\begin{table}[h]
	\centering
    \caption{Ablation studies on the M$^3$FD dataset. Bi-direction stands for using bi-direction decoupled focus on the backbone. DECA stands for using the DECA module. DEPA stands for using the DEPA module.}
	\begin{tabular}{@{}ccc|cc@{}}
		\toprule
		\makebox[0.15\textwidth]{Bi-direction} & \makebox[0.15\textwidth]{DECA} & \makebox[0.15\textwidth]{DEPA} & \makebox[0.15\textwidth]{mAP$_{50}$} & \makebox[0.15\textwidth]{mAP$_{50-95}$} \\ \midrule
		&      &      & 80.8  & 54.3     \\
		& \Checkmark    &      & 85    & 58.7     \\
		&      & \Checkmark    & 84.4  & 57.8     \\
		& \Checkmark    & \Checkmark    & 85.2  & 58.9     \\
		\Checkmark            & \Checkmark    & \Checkmark    & \textbf{86.6}  & \textbf{59.6}     \\ \bottomrule
	\end{tabular}
	\label{ablation modules}
\end{table}

Firstly, we verify the impact of the use of the bi-directional decoupled focus, DECA and DEPA modules on the model, respectively. The experimental results are shown in Table~\ref{ablation modules}. It can be seen that DECA and DEPA improve the detection accuracy of the model more obviously. The use of DECA and DEPA modules alone improves mAP$_{50}$ by 4.2$\%$ and 3.6$\%$, as well as mAP$_{50-95}$ by 4.4$\%$ and 3.5$\%$, compared to the baseline network trained merely by visible images. While the improvement of DECA is more obvious than that of DEPA. The joint use of them improves mAP$_{50}$ by 4.4\% and mAP$_{50-95}$ by 4.6\%, respectively.Moreover, the object detection accuracy is further improved using all three modules at the same time, with the two metrics improving by 5.8\% and 5.3\%, respectively.

In the DECA and DEPA modules, the channel weights and spatial pixel weights, which incorporate both semantic and spatial information from two modalities, are utilized to respectively enhance the semantic and structural information within the single-modality channel weights and spatial pixel weights. The enhanced weights are then applied to the single-modality feature maps to achieve dual enhancement. By fully leveraging the advantages of each modality and their complementary information within the feature space, the use of DECA and DEPA results in improving the performance of cross-modality object detection. Since we are utilizing deep features, each feature map contains stronger semantic information compared to spatial information. As a result, the enhancement effect of DECA on the model is more pronounced compared to that of DEPA.
% (定性分析)在DECA、DEPA两个模块作用时,它们利用融合了两个模态语义与空间信息的通道权重和空间像素权重,分别强化了单模态通道权重与空间像素权重中的语义与结构信息。再利用增强之后的权重分别作用于单模态特征图中,以达到双重增强的目的。由于在特征空间内充分利用了两模态各自的优势与它们之间的互补信息,DECA与DEPA被使用之后,模型跨模态目标检测的性能得到了提升。由于我们使用的都是深层特征,每张特征图中包含的语义信息要强于空间信息,所以导致DECA对模型的提升效果相较于DEPA更为明显。

Furthermore, in order to investigate how to make the dual enhancement mechanism in DECA and DEPA relieves the interference between two-modality images and obtain cross-modality channel weights and pixel weights better, we choose different hyperparameters in the feature mixing part in DEPA and cross-modality weight extraction part in DECA, respectively. 

\begin{table}[h]
	\centering
    \caption{Performance of different kernel sizes used in DEPA to get the mixed feature.}
	\begin{tabular}{@{}c|c|cc@{}}
		\toprule
		\makebox[0.15\textwidth]{Layer} & \makebox[0.2\textwidth]{Kernel Size} & \makebox[0.15\textwidth]{mAP$_{50}$} & \makebox[0.15\textwidth]{mAP$_{50-95}$} \\ \midrule
		\multirow{3}{*}{Conv} & 3 $\times$ 3 & $\mathbf{85.3}$  & $\mathbf{58.9}$     \\
		& 5 $\times$ 5 & 85.1  & 58.4     \\
		& 7 $\times$ 7 & 85.1  & 58.1     \\ \bottomrule
	\end{tabular}
	\label{ablation kernel size}
\end{table}

For DEPA, we use different convolution kernel sizes to get the spatial pixel weights of two modalities. The results are shown in Table~\ref{ablation kernel size}. We believe that as the convolution kernel size increases, more and more redundant information within each single modality is also integrated, thereby increasing mutual interference between the two modalities and hindering feature enhancement. It is found that for feature maps with different scales, when the number of convolutional layers is the same, the kernel size of 3×3 can better model the spatial pixel information.

% \vspace{-20pt}

\begin{table}[h]
	\centering
    \caption{Performance of different ways to generate $W_{Mix_0}$ through Cross-Modality Weight Extraction in DECA.}
	\begin{tabular}{@{}c|c|cc@{}}
		\toprule
		\makebox[0.15\textwidth]{Layer} & \makebox[0.25\textwidth]{Number of Layers} & \makebox[0.15\textwidth]{mAP$_{50}$} & \makebox[0.15\textwidth]{mAP$_{50-95}$} \\ \midrule
		\multirow{3}{*}{Conv} & 1 & \ding{55} & \ding{55} \\
		& 2 & 84.5  & $\mathbf{58.1}$     \\
		& 3 & $\mathbf{84.9}$  & 57.8     \\
		\midrule
		\multirow{2}{*}{\makecell[c]{Depth-wise \\ Conv}} & 2 & 84.5  & 58.3     \\
		& 3 & $\mathbf{85.2}$  & $\mathbf{58.9}$    \\ \bottomrule
	\end{tabular}
	\label{ablation layer type}
\end{table}

% 我们认为,随着卷积核的增大,单模态中冗余的信息也随之被融合,这样增加了两模态之间的相互干扰,不利于特征的增强。

For DECA, we try to use different types of convolutions with different numbers of layers for cross-modality channel weight extraction. The experiment results are shown in Table~\ref{ablation layer type}. We firstly attempt to directly extract the weights of each channel through one layer of convolution with the same size as the original feature map. However, we find that the model cannot converge if the layer number is set to 1. Then, we set the number of convolution layers to 2 and 3 successively, and find that the weights of each channel can be better extracted when it is 3. For channel weight extraction, we find that the depth-wise convolution~\cite{chollet:Xception} is more suitable for guiding the training process because of its fast convergence rate, which demonstrates its advantages. 

\begin{table*}[h]
	\centering
    \caption{Performance comparison with other detectors. Visible stands for training the model using visible images, infrared stands for training the model using infrared images. Cross-modality stands for using two-modality images for training.}
	\begin{tabular}{c|c|c|c}
		\toprule
		\makebox[0.35\textwidth]{Method}   & \makebox[0.25\textwidth]{Modality}    & \makebox[0.19\textwidth]{mAP$_{50}$}  & \makebox[0.19\textwidth]{mAP$_{50-95}$}  \\ \midrule
		\multirow{3}{*}{Swin Transformer~\cite{liu:Swin}} & visible        & 76.4  & 44.9     \\
		& infrared       & 72.6  & 41.9     \\
        & cross-modality & 73.8    &42.6        \\
		\multirow{3}{*}{CenterNet2~\cite{zhou:CenterNet2}}       & visible        & 78.5  & 52.4     \\
		& infrared       & 65.3  & 42.4     \\
        & cross-modality & 70.2    &46.5        \\
		\multirow{3}{*}{Sparse RCNN~\cite{sun:sparseRCNN}}      & visible        & 82.4  & 49.6     \\
		& infrared       & 76.4  & 44.8     \\
        & cross-modality & 78.2    &47.3        \\
		\multirow{3}{*}{YOLOv7-tiny~\cite{wang:YOLOV7}}      & visible        & 82.1  & 51.6     \\
		& infrared       & 78.1  & 48.4     \\
        & cross-modality & 80.1    &49.8        \\
		\multirow{3}{*}{YOLOv7~\cite{wang:YOLOV7}}           & visible        & 90.4  & 61.3     \\
		& infrared       & 87.9  & 58.3     \\ 
        & cross-modality & 88.3    &59.6        \\
		\multirow{3}{*}{YOLOv8n~\cite{YoloV8}}          & visible        & 80.8  & 54.3     \\
		& infrared       & 78.3  & 52.3     \\ 
        & cross-modality & 79.2    &52.8        \\
		\multirow{2}{*}{YOLOv8l~\cite{YoloV8}}          & visible        & 88.3  & 61.8     \\
		& infrared       & 86.5  & 59.6     \\ \midrule
		DEYOLO-n(ours)                         & Cross-modality & 86.6  & 58.9     \\
		DEYOLO-l(ours)                         & Cross-modality & $\mathbf{91.2}$  & $\mathbf{66.3}$    \\
		\bottomrule
	\end{tabular}
	\label{other detector}
\end{table*}

\subsection{Comparison with State-of-the-Arts models}

At last, we compare DEYOLO with recent state-of-the-art fusion models and object detection models on the M$^3$FD~\cite{liu:Tardal} and LLVIP~\cite{jia:LLVIP} datasets. Here we select YOLOv8-n and YOLOv8-l as our baseline.

As shown in Table~\ref{other detector}, due to utilization of different information from two modalities, DEYOLO outperforms all single-modality object detection models. In addition, mAPs of the detectors trained using visible images are higher than those of the detectors trained with infrared images. But none of the single-modality detectors can surpass DEYOLO, which uses the dual feature enhancement mechanism. Particularly, DEYOLO outperforms ViT-based models, such as Swin Transformer~\cite{liu:Swin} and Sparse RCNN~\cite{sun:sparseRCNN}. The ViT-based models only considers single-modality global correlation, while DEYOLO additionally uses the complementary information between two modalities extracted by DECA and DEPA without conflicts. % Compared to models based on self-attention mechanisms, such as Swin Transformer~\cite{liu:Swin}, .etc, model single-modality global correlation,  DEYOLO takes one step further, which utilities complements between two modalities by DECA and DEPA. DEYOLO introduces cross-modality information to guide the representation of features while diminishing the interference, takes full advantage of both modalities, and consequently improves the detection accuracy. % DEYOLO utilizes cross-modality information to guide each other by using the DECA and DEPA modules to take full advantage of both modalities . Therefore, the model is not only unaffected by the interference of the images of another modality when it is introduced, but also makes the model detection accuracy improve.

\begin{table*}[t]
	\centering
    \caption{Performance comparison with fusion-and-detection works.}
	\begin{tabular}{c|c|c|c|c}
		\toprule
		\makebox[0.19\textwidth]{Dataset} & \makebox[0.29\textwidth]{Method}  & \makebox[0.15\textwidth]{Modality} & \makebox[0.175\textwidth]{mAP$_{50}$} & \makebox[0.175\textwidth]{mAP$_{50-95}$} \\ \midrule
		\multirow{10}{*}{M$^3$FD~\cite{liu:Tardal}} & IRFS~\cite{WANG2023101828}  & \multirow{10}{*}{\shortstack{cross- \\ modality}}    &  81.2  & 55.8     \\
		& Tardal~\cite{liu:Tardal}   &       & 81.0    & 54.9     \\
		& CDDFuse~\cite{Zhao:CDDFuse} &        & 80.3  & 54.9     \\
		& PIAFusion~\cite{tang:PiaFusion} &     & 80.6  & 54.9     \\
		& Swin Fusion~\cite{ma:SwinFusion} &    & 80.2  & 54.7     \\
		& DetFusion~\cite{sun:DetFusion}    &   & 80.6  & 55.0       \\
		& SeAFusion~\cite{tang:SeAFusion}   &    & 80.7  & 55.4     \\
		& U2Fusion~\cite{xu:U2Fusion}     &   & 79.2  & 53.8     \\
		& DEYOLO-n(ours) & & 86.6  & 58.9     \\
		& DEYOLO-l(ours) & & $\mathbf{91.2}$  & $\mathbf{66.3}$     \\
		\midrule
		\multirow{9}{*}{LLVIP~\cite{jia:LLVIP}}&  IRFS~\cite{WANG2023101828} & \multirow{9}{*}{\shortstack{cross- \\ modality}}           & 94.0    & 60.7     \\
        & Tardal~\cite{liu:Tardal}    &       & 94.5  & 63.3       \\
        & CDDFuse~\cite{Zhao:CDDFuse}     &    & 92.1  & 57.5     \\
        & PIAFusion~\cite{tang:PiaFusion}    &     & 96.1  & 62.4     \\  
        & Swin Fusion~\cite{ma:SwinFusion}  &    & 93.3  & 59.4     \\
        & MFEIF~\cite{9349250}    &      & 95.8  & 64.0       \\
		& SeAFusion~\cite{tang:SeAFusion} &      & 96.2  & 64.0       \\
		& U2Fusion~\cite{xu:U2Fusion}   &     & 92.2  & 58.3     \\
		& DEYOLO-n(ours) & & $\mathbf{96.8}$  & $\mathbf{65.4}$    \\
		\bottomrule
	\end{tabular}
	\label{other fusion}
\end{table*}

It can be observed that some fusion-and-detect methods, such as DetFusion~\cite{sun:DetFusion} and U2Fusion~\cite{xu:U2Fusion}, as shown in Fig.~\ref{fig:comparison} (b) and (d), produce fused images which look more like the infrared images, lacking partial texture and color information required for detection tasks. On the other hand, the fused images obtained by the other methods including SeAFusion~\cite{tang:SeAFusion} and Tardal~\cite{liu:Tardal}, do not effectively capture rich structural information in the infrared image (e.g., Fig.~\ref{fig:comparison} (c)). The comparison methods fail to balance the texture and structure information of both modalities to improve the detection accuracy. In contrast, DEYOLO first exploits the advantages of both modalities through bi-direction decoupled focus and then utilizes the DECA and DEPA modules based on a dual-enhancement mechanism to reduce the mutual interference between the two modalities, thereby improving the detection accuracy.
% We observe that some of the fusion-and-detection methods in Table~\ref{other fusion} such as DetFusion~\cite{sun:DetFusion} and U2Fusion~\cite{xu:U2Fusion} output fusion images precisely matching the object poses of infared modality while the others including  Tardal~\cite{liu:Tardal} and SeAFusion~\cite{tang:SeAFusion} produce texture-rich results. However, none of them can keep a balance of maintaining information between two modalities as regards detection. Directly adopting the detection-relative objective, our method only  extract the most valuable and harmony features for improving detection accuracy.{figs/comparison.pdf}

\begin{figure}[H]
  \centering
  % 第一张图片
  \begin{minipage}{0.49\textwidth}
    \centering
    \includegraphics[width=\linewidth]{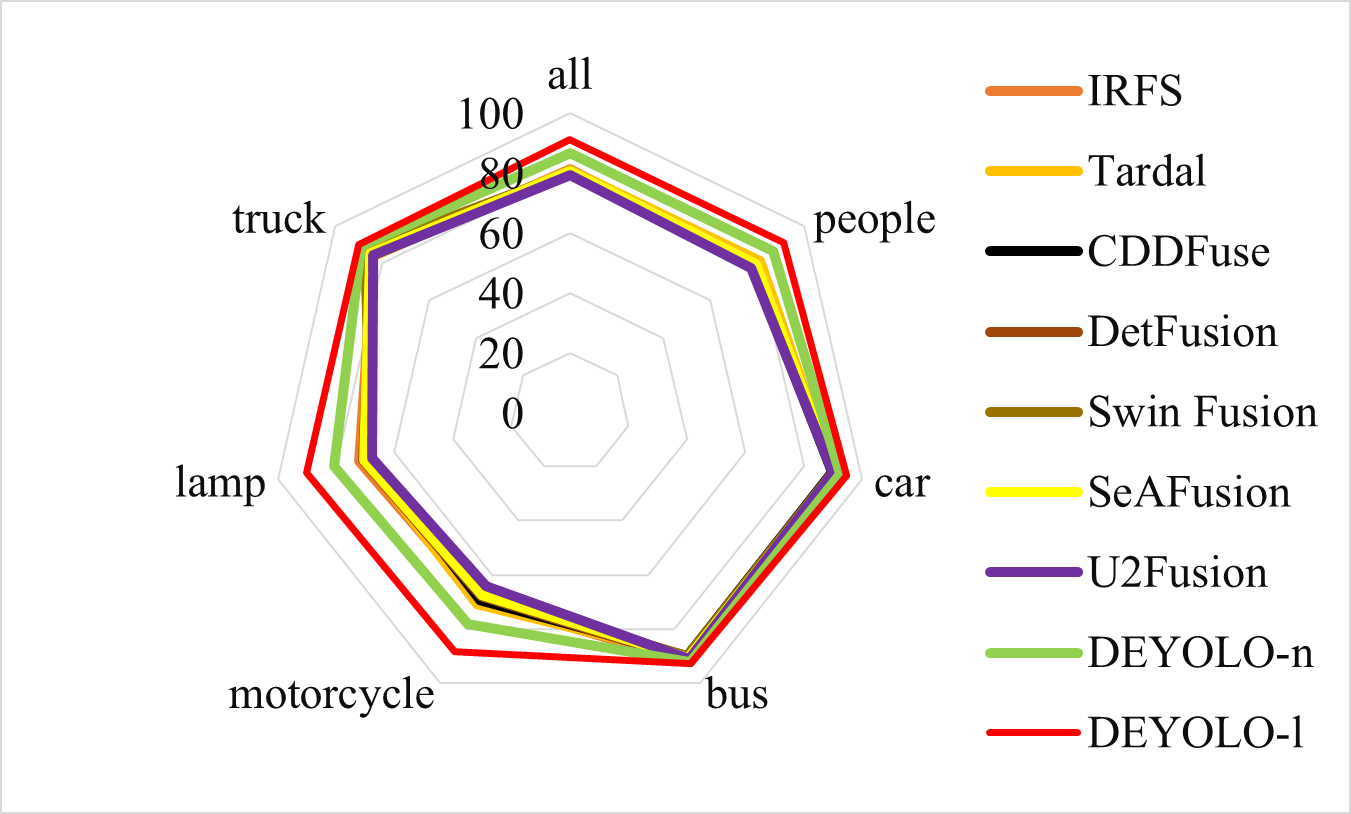}
    \vspace{-0.3cm}
    \caption{mAP$_{50}$ in specific categories}
    \label{map50 different classes}
  \end{minipage}
  \hfill
  % 第二张图片
  \begin{minipage}{0.49\textwidth}
    \centering
      \includegraphics[width=\linewidth]{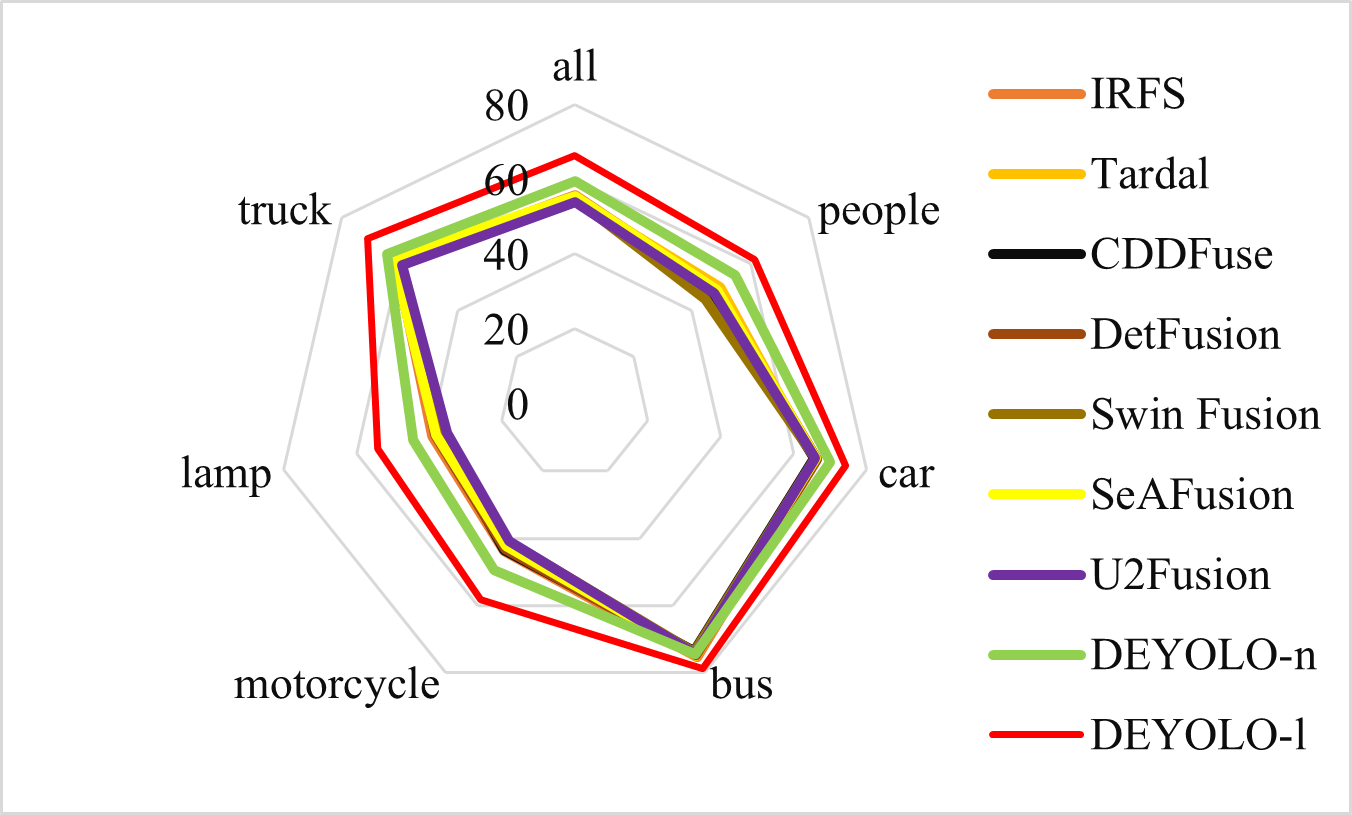}
    \vspace{-0.3cm}
    \caption{mAP$_{50-95}$ in specific categories}
    \label{map50-95 different classes}
  \end{minipage}
\end{figure}

As shown in Table~\ref{other fusion}, the performance of our method on both datasets is better than that of the state-of-the-art fusion-and-detection methods. Specifically, in M$^3$FD~\cite{liu:Tardal} dataset the mAP$_{50}$ and mAP$_{50-95}$ of DEYOLO-n are higher than those of the other models by 5.4\% and 3.1\% at least, respectively. And the improvement of the mAP$_{50}$ and mAP$_{50-95}$ of DEYOLO-l can reach more than 10.0\% and 10.5\%, respectively.  Meanwhile, in LLVIP~\cite{jia:LLVIP} dataset, we observe at least 0.6\% and 1.4 \% improvement on the mAP$_{50}$ and mAP$_{50-95}$ of DEYOLO-n, respectively. In addition, in Fig.~\ref{map50 different classes} and Fig.~\ref{map50-95 different classes}, the detection results of every category in M$^3$FD dataset also shows the superiority of our method. 
We have re-split the datasets into training, validation, and test sets in a 3:1:1 ratio. After dividing the test set as described above, the mAP$_{50}$ on the test/validation sets of the two datasets are 85.7\%/86.6\% and 96.4\%/96.8\%, respectively.

To validate the generalization ability of our model, experiments were conducted on the KAIST dataset, as shown in Table \ref{compar_kaist}. Unlike the M$^3$FD and LLVIP datasets, KAIST consists of pairs of RGB and thermal images. Thermal images, unlike infrared images studied in our research, exhibit lower imaging quality and significant differences. Therefore, these experiments serve as an extended validation of our model. From Table \ref{compar_kaist}, it is evident that our method does not achieve state-of-the-art (SOTA) performance but outperform the majority of existing methods.

\begin{table*}[h!]
\caption{Comparison with other RGB-T detectors on KAIST dataset.}
\centering
\setlength{\tabcolsep}{10pt}
    \begin{tabular}{c|ccc}%四个c代表该表一共四列,内容全部居中
    \toprule%第一道横线
    \textbf{Methods} & \textbf{ALL} & \textbf{Day} & \textbf{NIGHT}\\
    \midrule%第二道横线 
    RPN+BDT\cite{liu2016multispectral}                 & 29.83     & 30.51     & 27.62 \\
    TC-DET\cite{TC-DET}                 & 27.11     & 34.81     & 10.31 \\
    Halfway Fusion\cite{Halfway}         & 25.75     & 24.88     & 26.59 \\
    IATDNN\cite{IATDNN}                & 26.37     & 27.29     & 24.41 \\
    IAF R-CNN\cite{IAF}             & 20.59     & 21.85     & 18.96 \\
    CIAN\cite{CIAN}                        & 14.12     & 14.77     & 11.13 \\
    DEYOLO(ours)            & 15.45    & 17.23    & 12.23 \\
    \bottomrule
    \end{tabular}
    \label{compar_kaist}
\end{table*}

% We observed that although fusion-and-detection methods can produce fusion images containing details from both modalities, they aim at human perception rather than detection accuracy. Directly optimized by the YOLOv8 loss, our method only calculates task-relevant gradient and doesn't extract redundance features for improving detection accuracy.
% Extensive research has shown that images that are easy for humans to understand often contain a lot of redundant information for detectors. These redundant information will not only bring computational burden to the detector, but also could be misleading to the detector. The fusion-and-detection models focus on generating images cater to human visual preference, wasting computational resources on learning information that the detector does not need. Whereas DEYOLO gains the gradients needed by the detector, improves the detection accuracy.
% Besides, we visualize our method and sever other fusion-and-detection work in Fig.~\ref{fig:comparison} which shows the high robustness and performances.

% \vspace{-5pt}
\section {Conclusion}
In this paper, we propose DEYOLO using the dual enhancement mechanism for cross-modality object detection in complex-illumination environments. DECA and DEPA are designed to fuse the feature maps of two modalities between the backbone and the detection heads. And the bi-direction decoupled focus is proposed in the backbone to improve the feature extraction capability. The superiority of this method is verified on two datasets. It is worthwhile to point out that, both DECA and DEPA proposed in this paper can be used as a plug-and-play module for wider applications in other models to solve the problem of object detection in complex environments. And this will be the topic in our future work.

\section {Acknowledgement}
This work was supported in part by the Natural Science Foundation of China under Grant U21A20486, 62473208 and 62401294, in part by the Tianjin Science Fund for Distinguished Young Scholars under Grant 20JCJQJC00140, in part by the major basic research projects of the Natural Science Foundation of Shandong Province under Grant ZR2019ZD07, in part by the Postdoctoral Fellowship Program of CPSF under Grant GZC20240753, and in part by the Fundamental Research Funds for the Central Universities under Grant 078-63243158.
% This work was supported in part by the Natural Science Foundation of China under Grant U21A20486 and 62073178, in part by the Tianjin Science Fund for Distinguished Young Scholars under Grant 20JCJQJC00140, in part by the major basic research projects of the Natural Science Foundation of Shandong Province under Grant ZR2019ZD07, in part by the Postdoctoral Fellowship Program of CPSF under Grant Number GZC20240753.

\bibliographystyle{splncs04}
\bibliography{icpr24}

\end{document}